\documentclass[letterpaper]{article}
\usepackage{aaai25}
\usepackage{times}
\usepackage{helvet} 
\usepackage{courier}
\usepackage[hyphens]{url} 
\usepackage{graphicx}
\usepackage{natbib}
\usepackage{caption} 
\usepackage{bbold}
\usepackage[T1]{fontenc}
\usepackage{import}
\usepackage{diagbox}
\usepackage{lipsum}
\usepackage{placeins}
\usepackage{algorithm}
\usepackage{algorithmic}
\usepackage{amsmath}
\usepackage{amsthm}
\usepackage{dsfont}
\usepackage{array}
\usepackage{shortcuts}
\usepackage{tabularx,ragged2e} 
\newcolumntype{C}{>{\Centering\arraybackslash}X}
\usepackage{mathtools}
\usepackage{nccmath}
\usepackage{makecell}
\usepackage{relsize,etoolbox}
\AtBeginEnvironment{quote}{\smaller}
\usepackage[dvipsnames]{xcolor}
\usepackage{colortbl}
\definecolor{strongrep}{rgb}{.9,.2,.1}
\definecolor{strongdem}{rgb}{.1,.2,.9}
\definecolor{modrep}{rgb}{.7,.2,.1}
\definecolor{moddem}{rgb}{.1,.2,.7}

\usepackage{newfloat}
\usepackage{listings}
\DeclareCaptionStyle{ruled}{labelfont=normalfont,labelsep=colon,strut=off} 
\floatstyle{ruled}
\newfloat{listing}{tb}{lst}{}
\floatname{listing}{Listing}

\title{\textsc{PublicSpeak:} Hearing the Public with a Probabilistic Framework}
\author {
    Tianliang Xu\textsuperscript{\rm 1},
    Eva Maxfield Brown\textsuperscript{\rm 2},
    Dustin Dwyer\textsuperscript{\rm 3},
    Sabina Tomkins\textsuperscript{\rm 1}
}
\affiliations {
    \textsuperscript{\rm 1} University of Michigan School of Information\\
    \textsuperscript{\rm 2}University of Washington Information School\\
    \textsuperscript{\rm 3}Michigan Public\\
    tianlix@umich.edu, 
    evamxb@uw.edu, 
    dustin@michiganpublic.org,
    stomkins@umich.edu
}

\graphicspath{figures}

\usepackage{bibentry}

\begin{document}

\maketitle

\begin{abstract}
    Local governments around the world are making consequential decisions on behalf of their constituents, and these constituents are responding with requests, advice, and assessments of their officials at public meetings. So many small meetings cannot be covered by traditional newsrooms at scale. We propose \ourmethod, a probabilistic framework which can utilize meeting structure, domain knowledge, and linguistic information to discover public remarks in local government meetings. We then use our approach to inspect the issues raised by constituents in 7 cities across the United States. We evaluate our approach on a novel dataset of local government meetings and find that \ourmethod{} improves over state-of-the-art by 10\% on average, and by up to 40\%. 
\end{abstract}

 \begin{links}
 \link{Code}{https://github.com/politechlab/publicspeak}
 \end{links}
\section{Introduction}
Communities across the United States are grappling with complex and even existential issues, such as
housing, policing, pollution, and the (under)provision of public services \citep{anzia2022local}. 
Many of the policies that
address these issues
are formed at the local level. To effectively participate in local governance, community members need political information to have knowledge of current issues and policies, as well as
about political procedures, norms and institutions.

The role of informing communities about local politics, especially those outside of major metropolitan areas, has traditionally fallen to local news 
\citep{powe1992fourth,hanitzsch2018journalism}.
However, local news has been in decline \citep{sullivan2020ghosting} and 
scholars have questioned whether the information  provided by traditional outlets is truly relevant for local contexts and  political participation \citep{usher2017appropriation,wilner2021me}.
A contemporary model of local information infrastructure
will likely depart from printed media \citep{george2008internet,newman2020reuters,parasie2013data}.  Content curated for particular communities may enable more profound action, accountability, and inclusivity \citep{velazquez2014solidarity}. In the face of declining budgets and shifting interests, there is a need for innovative  approaches to local news. 

This work is a joint effort by a reporting team and data scientists to use artificial intelligence (AI) to support existing news architecture. We leverage the fact that local governments are increasingly publishing video recordings of their  meetings online \citep{piotrowski2010analytic}. 
The ability to convert video recordings of local government meetings to reliable text-data presents an opportunity to assist local journalists as well as citizens. A number of researchers and organizations have begun to generate text from local government meetings.\footnote{ The Council Data Project, Local Lens, and Search Minutes
}
Yet, so far, none of these organizations have extracted the remarks made by the public, a crucial need for local reporters. To address this gap, we propose \ourmethod{}, a probabilistic framework for extracting public remarks. Here, we evaluate our approach on 30,000 remarks across seven cities in the United States.

\subsection{Hearing From the Public, a Data-Driven Opportunity}
Public digital postings of city council meetings offer a rich opportunity to understand the concerns of the public.
As of 2022, more than 2,800 local government units 
had broadcast meeting videos on YouTube alone \citep{barari2023localview}
and  a plenitude of meetings have been mined from other channels \citep{maxfield2022councils}.
The increase in recordings of public meetings 
has co-occurred with
improved performance of automatic speech recognition (ASR) algorithms.
We address the critical gap in existing approaches for mining this data.

Public remarks at local government meetings provide a unique window into the priorities of community members. While cities, townships and school boards are required to post agendas of meetings beforehand, signifying the board members’ priorities, public comments can offer residents an open forum. The comments represent the priorities of the public, rather than the priorities of elected officials. Community members'  public remarks frequently guide news coverage of local government.
As local news outlets have cut staff, fewer journalists are able to monitor these comments, and no equivalent source of information on community priorities has emerged. Our work addresses that need by 
making the following \textbf{contributions}:

\begin{itemize}
   
    \item \ourmethod, a probabilistic  framework for discovering public remarks. 
    \item An evaluation of this framework across seven cities with differing populations and priorities. We demonstrate that our approach consistently outperforms state-of-the-art, by up to 40\% in some cities. 
    \item A demonstration of how this framework can be used to discover common discussion topics amongst the public. 
\end{itemize}

The priorities expressed by the public at city council meetings often precede government action at the local, state and national level. For example, the calls for reform around policing from local activists have affected local policing policies \citep{brown2021philadelphia,novick2022black}, and spurred the creation of offices of violence prevention \citep{phelps2021police}. 
These calls, though made locally, often grow to national significance. And a computational framework that allows for extracting the full text of public remarks from one local meeting 
could be scaled to extract from thousands of meetings nationwide. This would allow for the creation of an entirely new data set of political priorities in the U.S. and around the world, from the ground up. Towards this end we have also developed a pipeline 
for automatically collecting public meeting records from cities across the United States. 
Our intention is to continuously publish datasets of 
public remarks for community organizers, journalists, and researchers.

\section{Public Remarks}
Our goal is to categorize remarks made at local government meetings. 
Meeting structure varies considerably across municipalities. 
Even within the same state, municipalities may employ different structures. 
Each municipality decides when the public is invited to speak, 
and the topics which they are 
sanctioned to speak about.

Across cities, council meetings commonly 
host two periods during which the public can speak:
public comments and public hearings. 
Generally, during public comments, participants may speak on any issue they wish, as long as they remain within the time limit. 
During public hearings, the public may only remark on the issue discussed during the hearing. 
While some cities restrict public comments to agenda items, and may introduce 
other rules to control the meeting \citep{vpm2024}, 
in general public comments are less restricted than hearings
and provide a better view into the concerns of the public. 
They are also more ubiquitous across cities. 
Consequently, 
we  categorize remarks made during a public comment period 
outside of public hearings. 

\subsection{Problem Definition}
Our goal is to 
solve the binary 
classification task of identifying all public comments.
Let $\u$ refer to  an utterance made during  a public meeting. We say that $\u$ is a public remark only if it is made by a meeting participant assuming the role of \textit{a member of the public} and uttered either during \comments{} or \hearings. 
If $\u$ occurs during \comments{} we call it a public comment, $\pc$.

Given a dataset of $\mathds{D}= \{(x_1,y_1),\dots, (x_N,y_N)\}$
 where $y_i$ is a remark category $\in \{ \pc, Other\}$, and 
 $x_i$ belongs to a multimodal feature space $\mathcal{X}$ which depends on the available data, 
 our goal is to
 learn a function $f: \mathcal{X} \to \{ \pc, Other\}$. 
 
Along with remark categories, we jointly infer  discussion sections and speaker roles. 
 However, our goal is not necessarily to have a full 
 accounting of  meeting discussion sections and speaker roles, 
but rather to produce a final dataset of public comments. 

\section{Our Approach}
\label{sec:approach}
Our approach, \ourmethod{}, uses the structure of governmental meetings to extract the comments made by the public. 
Public remarks are linguistically distinct from those made by members of government. However, when it comes to disambiguating remarks made by the public at different points in a meeting, linguistic signals can be misleading. 
For example, the style of a remark made in  \hearings{} may be similar to that of one made in \comments. 
Our approach allows us to fuse knowledge of when different remarks should be made (structural signals), with 
 information about how those remarks are made (linguistic signals). 

Additionally, to disambiguate remarks made during \hearings{} from those made in \comments,
we 
reframe the learning problem as a multi-class classification task. 
In this setting,
if $\u$ occurs during \hearings{} we call it $\ph$, and 
 $y_i$ is a remark category $\in \{\pc, \ph, Other\}$.
 Here, our goal is to learn a function 
  $f: \mathcal{X} \to \{ \pc, Other,\ph\}$.
We can then form a dataset from the utterances classified as $\pc$. 
We expect this approach to perform the best when there are 
many \hearings{} in a meeting, and when public remarks
in \hearings{}
closely resemble those in \comments.

We construct our model with Probabilistic Soft Logic (PSL) \citep{bach2017hinge}, an opensource probabilistic programming framework which is well suited for this task. With PSL, we can 1) jointly infer meeting structure, speaker roles, and comment types, 2) exploit relationships between these entities, and 3) inform our predictions with domain knowledge without resorting to brittle heuristics. For example, we can encode that domain knowledge that members of the public tend to have fewer remarks of greater duration than city council member, without enforcing that this always be the case.

\subsection{Probabilistic Soft Logic}
In PSL, relationships between variables are encoded with weighted
logical rules. These rules can capture dependencies not only from
observed features to target variables, but between target variables. This expressivity allows us to encode both domain knowledge about how meetings are run, and relationships between utterances made within a meeting.

Here, we use the fact that both a $\ph$ and a $\pc$ need to occur within specific sections of the meeting. To do so, we introduce the predicate $\textsc{Section}$, which takes three arguments: a meeting id $M$,  an utterance id $U$, 
and one of the following three section types: \comments, \hearings, or \textsl{Other}. \textsl{Other} contains all meeting sections which are not open to public comment, for example roll call and budget discussions. Knowledge of one's role within a meeting can also be used to infer whether one is making a public remark. For example, a project applicant who is explaining a proposed building project is not acting as a member of the public when they speak. Whenever they make a remark during a public hearing it is within their role as a project applicant, generally to answer questions or provide information about the building project. In practice, speakers in government meetings assume many different roles. We model whether they are acting as a member of the public or not. We do this with the predicate $\textsc{SpeakerRole}$ which takes three arguments: a meeting id $M$, a speaker id $S$, and a role which is either \textit{Public}, or \textit{Other}.

Finally, our goal is to use meeting structure and speaker roles to discover the remarks which are made by or on behalf of members of the public, during public comments and hearings. 
We model these remarks with the predicate \textsc{Remarktype}, which takes three arguments: $M$, $U$, and $T$, which correspond to a meeting id, an utterance id, and a remark type.
In the binary specification $T$ can be (either $\pc$, or \textsl{Other}), while in the multi-class setting it can also be $\ph$. 

A predicate and its arguments form a logical atom, which can assume soft truth values in [0, 1]. With these predicates, and a weight $w_{str}$ that reflects the relative importance of this rule, we define a rule which expresses the relationship between sections, speaker roles, and remark types:
\begingroup\makeatletter\def\f@size{7}\check@mathfonts
\def\maketag@@@#1{\hbox{\m@th\large\normalfont#1}}%
\begin{align*}
w_{str}: \textsc{Section}(M,U,X) \wedge \textsc{Spoken}(M,U,S) \wedge\textsc{SpeakerRole}(M,S,Y) & \\
\Longrightarrow\textsc{Remarktype}(M,U,X).&
\end{align*}
\endgroup
Combined with data, a PSL model defines a joint probability distribution over sections, speaker roles, and comment types. This distribution is expressed
with a hinge-loss Markov random field (HL-MRF) \citep{bach2017hinge}, a general class
of conditional, continuous probabilistic graphical models which provide the advantage of both high efficiency and expressivity.  
We use PSL to infer 
remark types by
fusing structural information with 
linguistic content from the utterances themselves.

\subsection{\ourmethod{}}
We model the fact that the most frequent speaker role 
is \textit{Other} with the first rule in \tabref{priors}.
We additionally constrain that
a single utterance belongs to one meeting section, 
that each utterance has a single label, 
and that
speakers 
have only one role within a meeting. 
The rules discussed below are unchanged depending on whether 
we frame the problem with two classes or three. 
The only difference is in the values which the 
argument $T$ can take in the predicate $\textsc{Remarktype}$. 
As structure is important for identifying $\pc$, 
even in the binary setting, we still identify all meeting sections
and speaker roles independent of the task specification. 

\begin{table}[h!]
\centering
\resizebox{.68\columnwidth}{!}{%
\begin{tabular}{|ll|}
\hline
$w_{np}:$ &$\neg \textsc{SpeakerRole}(M,S,Public)$\\
$\infty:$&$\sum_{t} \textsc{Section}(M,U,t)$ = 1\\
$\infty:$&$\sum_{t} \textsc{Remarktype}(M,U,t)$ = 1\\
$\infty:$&$\sum_{t} \textsc{SpeakerRole}(M,S,t)$ = 1\\
    \hline
    \end{tabular}
    }
    \caption{ 
   Priors allow us to encode which categories we expect to be most prevalent.
    These are especially important, as we see class imbalance in all of the cities in the dataset. Rules with an infinite weight ($\infty$) are hard constraints.}
    \label{priors}
\end{table}

\subsubsection*{Meeting Structure}
The first rule in \tabref{structure} encodes the domain knowledge that meetings rarely begin directly with a public remark. Instead, the first utterance is typically an introduction to the meeting made by a government representative. 
To model consistency, we introduce the atom $\text{Precedes}(M,U_1,U_2)$, which is true if $U_1$ is the utterance immediately preceding $U_2$ and false otherwise. Here we show the template for this rule, while in the model we instantiate one rule for each combination of $X$ and $Y$. This allows us to learn a transition probability for each of the discussion sections. Next, we learn transition probabilities between different remark types within the same section. For example, in some localities a meeting manager offers a transition between each public speaker by saying something like \textit{Thank you. Our next speaker is Jane Doe}.

\begin{table*}[h!]
\centering
\resizebox{.68\textwidth}{!}{%
\begin{tabular}{|llcr|}
\hline
$w_{prior}:$ &$\textsc{First}(M,U)$& $\Longrightarrow$ & $\textsc{Section}(M,U,Other)$\\
$w_{t_{xy}}:$&$\textsc{Section}(M,U_1,X) \wedge \textsc{Precedes}(M,U_1,U_2)$& $\Longrightarrow$ & $\textsc{Section}(M,U_2,Y)$\\
$w_{a_{xy}}:$&$\textsc{Section}(M,U_1,X) \wedge \textsc{Remarktype}(M,U_1,X)$ $\wedge $&  &\\
&$\textsc{Precedes}(M,U_1,U_2)$& $\Longrightarrow$ & $\textsc{Remarktype}(M,U_2,Y)$\\
\hline
    \end{tabular}
    }
    \caption{Rules which encode meeting structure. \label{structure}}
\end{table*}

\subsubsection*{Speaker Roles}
Officials who are responsible for running a government meeting will speak quite often. In contrast, those making public remarks will speak less often, but their utterances tend to have greater length. 
We reflect these three observations with the first three rules in \tabref{speakerroles}. The predicate $\textsc{LongUtteranceRatio}_{\delta}$, captures the proportion of a given speaker's utterances which exceed 
a threshold $\delta$ of words. 
The final rule in \tabref{speakerroles} allows us
to learn the extent to which speakers 
assume different roles in each discussion section. 
For example, this rule allows us to model the fact
that public remarks only rarely occur during any section 
other than public comment or public hearings.

\begin{table*}[h!]
\centering
\resizebox{.68\textwidth}{!}{%
\begin{tabular}{|llcr|}
\hline
$w_{hc}:$&$\textsc{SpeaksOften}(M,S) $& $\Longrightarrow$ & $\textsc{SpeakerRole}(M,Other)$\\
$w_{lc}:$&$\textsc{SpeaksRarely}(M,S) $& $\Longrightarrow$ & $\textsc{SpeakerRole}(M,Public)$\\
$w_{lur}:$&$\textsc{LongUtteranceRatio}_{\delta}(M,S) $& $\Longrightarrow$ & $\textsc{SpeakerRole}(M,Public)$\\
$w_{c2sp_{xy}}$:& $\textsc{Section}(M,U,X) \wedge \textsc{Spoken}(M,U,S) \wedge$&  &\\
 &$\textsc{SpeakerRole}(M,S,Y)$&$\Longrightarrow$&$\textsc{Remarktype}(M,U,X)$\\
 
    \hline
    \end{tabular}
    }
    \caption{Rules which encode information about speaker roles.\label{speakerroles}}
\end{table*}

\subsubsection*{Linguistic Signals}
There is a rich amount of linguistic information within government meetings. 
We find that public comments and hearings 
are almost always introduced with some phrase which is standard 
for a given locality and meeting type. However, these phrases are not standard across localities, 
or even within localities across meeting types. 
For example, in one locality the beginning of the public comment  discussion section 
may be triggered with the phrase \textit{we now come to public comment reserve time}, 
while in another locality the phrase may be \textit{we will now hear from the citizens}. 
Similar phrases typically mark the beginning of public hearings. 
Unfortunately, there is much less structure with ending discussion sections. Often,
the next section will be signaled without the current session being explicitly ended. 
Thus, we only model the beginning of a given section, rather than its end. 
We do so with the predicates, \textsc{CommentTransition} and \textsc{HearingTransition}
which are true for an utterance $U$ in meeting $M$ if $U$ matches a locale-specific pattern
which typically indicates the beginning of \comments{} or \hearings. 
The last rule encodes the domain knowledge that members of the public tend to introduce themselves
before speaking, where the predicate  \textsc{Introduction}, is true for the utterance $U$ in meeting $M$ if $U$ 
is an introduction.  

\begin{table*}[h!]
\centering
\resizebox{.68\textwidth}{!}{%
\begin{tabular}{|lcr|}
\hline
$w_{cmt}:\textsc{CommentTransition}(M,U) $& $\Longrightarrow$ & $\textsc{Section}(M,U,``PC")$\\
$w_{hear}:\textsc{HearingTransition}(M,U) $& $\Longrightarrow$ & $\textsc{Section}(M,U,``PH")$\\
$w_{into}:\textsc{Introduction}(M,U) \wedge \textsc{Spoken}(M,U,S)$& $\Longrightarrow$ & $\textsc{RemarkType}(M,U,``PC")$\\
    \hline
    \end{tabular}
    }
    \caption{Rules for linguistic transition signals.\label{Linguistic rules}}
\end{table*}

\subsubsection*{Combining and Correcting AI Signals}
Finally, we incorporate information from state-of-the-art AI models. These models allow us to leverage the knowledge from extensive existing datasets.
For example, we  use generative AI (GenAI) to discover section transitions with meeting transcripts. We then learn the extent to which these unsupervised predictions are correct with the first rule in \tabref{airules}. Here the value of \textsc{SectionGenAI} for a particular meeting $M$, utterance $U$ and section type $X$ is the output of a conversational agent.  
Similarly, in addition to the linguistic cues mentioned above
we \textit{fine-tune} pre-trained language models (PLMs) to better utilize 
the linguistic information in each utterance (see the second rule). However these text-based predictions from the PLM
 ignore meeting structure, while the GenAI overly depends on it. 
 To address these inconsistencies
 we introduce the final rule.
 Note, in the binary task specification there are no rules for
 when the remark type may be $\ph$.

\begin{table*}[h!]
\centering
\resizebox{.68\textwidth}{!}{%
\begin{tabular}{|lcr|}
\hline
$w_{gai_x}:\textsc{SectionGenAI}(M,U,X) $& $\Longrightarrow$ & $\textsc{Section}(M,S,X)$\\
$w_{plm_x}:\textsc{RemarkTypePLM}(M,U,X) $& $\Longrightarrow$ & $\textsc{Remarktype}(M,U,X)$\\
$w_{fix_{xy}}:\textsc{RemarkTypePLM}(M,U,Y) \wedge \textsc{SectionGenAI}(M,U,X) $& $\Longrightarrow$ & $\textsc{Remarktype}(M,U,X)$\\
    \hline
    \end{tabular}
    }
        \caption{Rules which incorporate and correct AI Signals.  \label{airules}}
\end{table*}

\section{Empirical Evaluation}
We assess our framework on its ability to discover the public remarks 
in all city council meetings for 
seven cities across four states for roughly the duration of one year; which were selected in partnership with a non-profit news organization.   
Our goal was to cover a time frame sufficiently long 
so that we would not be susceptible to learning short-term 
patterns (such as memorizing phrases which are relevant to a particular event).

\subsection{Dataset}
Our goal is to have a scalable pipeline
to be able to continuously produce open-source
records of public remarks at local government meetings
across the United States. 
This pipeline consists of:
1) collecting, 2) processing, 3) annotating, 
and 4) publishing meeting records. 
Here, our focus is on using AI to 
produce a high-quality dataset of public remarks. 

\subsubsection*{Initial City Set}
To evaluate the ability of our method to scale to 
the United States, here we begin 
with cities which span four states 
and which represent a variety of council meeting structures, 
as well as socio-economic diversity. 
Towards this end, we 
include three cities from across the United States: 
Oakland, California; Richmond, Virginia; and Seattle, Washington.
We additionally include four cities in the state of Michigan: 
Ann Arbor, Jackson, Lansing, and Royal Oak. 
By collecting four cities within a single state we can 
better understand state-specific factors of participation, 
while contrasting these to other cities across the US 
allows us to evaluate the ability of our approach to generalize. 

\begin{table*}[]
\centering
\resizebox{0.99\textwidth}{!}{%
\begin{tabular}{|c|c|c|c|c|c|c|c|c|c|c|}
\hline
 & Seattle, WA & Oakland, CA & Richmond, VA & Ann Arbor, MI & Lansing, MI & Royal Oak, MI &Jackson, MI   \\ \hline
Population &737,015&440,646 & 226,610& 123,851 & 112,644 &58,211& 31,309    \\ \hline
Median Income & \$115,000 & \$93,000 & \$59,000 & \$78,000 & \$49,000 & \$93,000 & \$42,000 \\
\hline
Percent Asian & 17\% & 16\% &3\% & 16\% & 4\% &3\% &1\% \\
Percent Black &7\% & 21\% &40\% &7\% & 23\% &4\% &19\% \\
Percent Hispanic & 8\% & 29\% &11\% & 5\% & 14\% & 4\% &7\% \\
Percent white & 59\% & 30\% &43\% & 68\% & 55\% &86\% &65\% \\
\hline
\makecell{Percent bachelor's \\ degree or higher}  & 67\% & 51\% & 46\% & 79\% & 30\% & 61\% & 15\%  \\
\hline
    \end{tabular}%
    }
    \caption{Socio-economic characteristics of each city. \label{tab:demos}}
\end{table*}

We see in \tabref{tab:demos}
that these cities offer a range of socio-economic profiles. 
For example, Seattle is 
one of the 100 largest cities 
in the United States, while 
Jackson is relatively small. 
These cities also reflect a range of median incomes and education levels. 
The gap between the percentage of adults with a bachelor's degree or higher 
is particularly stark between Ann Arbor and Jackson, MI, for example. 
Finally, when it comes to race, these cities offer a rich canvas to test our approach. 
For example, Oakland, CA hosts significant populations of Asian, Black, 
and Hispanic residents. As a method which utilizes AI may suffer from bias for diverse populations, 
we decided it was imperative that our initial set of cities were heterogeneous in terms of race.

\subsubsection*{	Creating Transcripts to Annotate}
First, we collect video recordings (as .mp4 files)
of all city council meetings held in either the year 2021 or 2023 for seven cities. 
For the purpose of assessing feasibility, we 
selected  cities from that set which 
regularly publish videos either to YouTube 
or their government affiliated website
and which we anticipated would be accessible in the future.

Next, we transcribe these city council meetings with WhisperX \citep{bain2022whisperx}. We first use the Voice Activity Detection (VAD) model from Whisper \citep{whisper} to pre-segment the original audio files, ensuring that the boundaries of each segment do not fall within areas with vocal sounds. We then merge the original segments into approximately 30-second clips and use Whisper's Automatic Speech Recognition (ASR) model to transcribe each audio segment. Finally, the speaker IDs are inferred and assigned to each segment by the Speaker Diarization model \texttt{pyannote} \citep{pyannote1}.

Next, after inferring speaker IDs we group subsequent text snippets
made by the same speaker together into utterance groups.
This provides a final table  
where each line corresponds to a single utterance made by a disambiguated speaker. Note, this process does not produce speaker names or other descriptive characteristics, it simply disambiguates the speakers from each other.

\subsubsection*{Annotating transcripts}
We include the full schema description and an overview of the instructions for annotators in the supplementary material. 
To provide a brief overview, we asked annotators to determine three categorizations for each utterance: whether it was a transition (either implicit or explicit), the meeting section it occurred during, and the role of the speaker of the utterance. The transitions mark the beginning and end of \comments{} and \hearings{} and are redundant given the sections but decreased the difficulty of the task, as annotators were able to more efficiently identify sections after finding the transitions into and out of them. 
While in \ourmethod, we only consider two speaker roles, \textit{Public} and \textit{Other}, annotators further differentiated when members of government were speaking. 
To classify each remark we used the following rule: 
\[\medmath{
    y_i=  
\begin{dcases}
   \pc& \text{if } Role(U_i)=\textit{Public} \wedge Section(U_i)= \comments\\
    \ph & \text{if }  Role(U_i)=\textit{Public} \wedge Section(U_i)= \hearings \\
        \textit{Other},              & \text{otherwise}
\end{dcases}
}
\]

where $Role(U_i)$ is the role of the speaker of utterance $U_i$ and $Section(U_i)$ is the meeting section during which $U_i$ occurs. Here, the label $\ph$ is only present in the 3-class task specification. 
 Each meeting was reviewed by three annotators and we show the inter-annotator agreement in \tabref{tab:anno}.

\subsubsection*{Data Limitations}
Each module in our pipeline is susceptible to error. 
At the onset, videos have widely varying quality. 
Speakers fail to turn on their microphones, 
or an audience may be so loud that it is difficult to hear 
the assigned speakers. 
There are also errors which arise from the transcription 
and speaker identification algorithms. 
For example, we still find minor hallucinations in the transcripts. 
There are also minor errors in the transcripts
which can diminish the quality in the final dataset.
Finally, there are known racial disparities in ASR technologies which may differentially influence the cities in our dataset \citep{koenecke2020racial}.

We experimented with techniques for mitigating these errors. Regarding hallucinations, we observed the general characteristics of hallucinations. In order to minimize alterations to the original data as much as possible, we set up a simple filter to remove some obvious hallucinations, such as ``www.'', ``openai'', ``https'', etc. Initially, we segmented the original audio files based on different speakers. We found that this approach often resulted in many very short segments, around 1 second in length, where hallucinations frequently occurred. Consequently, we changed the order of the pipeline, first segmenting the original audio with WhisperX, then cutting and merging the segments into larger approximately 30-second segments before beginning transcription. This effectively reduced the occurrence of hallucinations.

\subsection{Quantitative Evaluation}
We assess our framework on its ability to discover the public comments in all city council meetings for seven cities for a held out period 
following the final meeting in the training data. We compare our approach to two custom baselines we develop, and to data-driven and state-of-the-art  models.

\begin{table*}[ht]

\centering
\fontsize{10pt}{12pt}\selectfont
\resizebox{0.99\textwidth}{!}{%
\begin{tabular}{|l|c|c|c|c|c|c|c|c|c|c|}
\hline
 &Seattle&Oakland&Richmond& Ann Arbor & Royal Oak & Jackson & Lansing   \\ \hline
Krippendorff's $\alpha$ & 0.982  & 0.900  & 0.831  & 0.918  & 0.931  & 0.876 & 0.953  \\ \hline
\makecell[l]{Agreement \\(\% of time all annotators agree)} & 0.993  & 0.969  & 0.971 & 0.980  & 0.989  & 0.981 & 0.988  \\ \hline
\makecell[l]{Percent of remarks \\which are public comments}& 15.20\% & 9.11\%  & 1.14\% & 7.04\%  & 5.76\%  & 5.90\% & 4.28\%\\
\hline
\makecell[l]{\textbf{Training} data 
date range}& Jan-Jul 2021  & Jan-Jul 2021  & Jan-Jul 2021 & Jan-May 2023  & Jan-May 2023  & Jan-May 2023 & Jan-May 2023\\
\makecell[l]{\textbf{Validation} data
date range}& Jul-Aug 2021   & Sep-Oct 2021  & Sep 2021 & Jun-Jul 2023  & Jun-Jul 2023  & Jun-Jul 2023 & Jun-Jul 2023\\
\makecell[l]{\textbf{Test} data
date range}& Oct 2021   & Nov 2021  & Oct-Dec 2021 & Aug-Sep 2023  & Aug-Sep 2023  & Aug-Sep 2023 & Aug-Sep 2023 \\
\hline
\makecell[l]{Total remarks overall} & 2481  & 8156  & 4124 & 3123  & 3788  & 4137 & 4253 \\
\hline
    \end{tabular}%
    }
\caption{Characteristics of the training data. The combined dataset contains a total of $\sim$30K remarks.
}
\label{tab:anno}
\end{table*}

\begin{table*}[h!]
\resizebox{\textwidth}{!}{%
\begin{tabular}
{|c|r|r|r|r|r|r|r|r|r|r|r|r|r|r|r|r|r|r|r|r|r|r|r|r|r|}
\hline
       \multicolumn{1}{|c|}{Model} &  \multicolumn{2}{|c|}{Seattle $\pc$-F1} & \multicolumn{2}{|c|}{Oakland $\pc$-F1} &   \multicolumn{2}{|c|}{Richmond $\pc$-F1}&
      \multicolumn{2}{|c|}{Ann Arbor $\pc$-F1}&
       \multicolumn{2}{|c|}{Lansing $\pc$-F1}&
       \multicolumn{2}{|c|}{Royal Oak $\pc$-F1}&
       \multicolumn{2}{|c|}{Jackson $\pc$-F1} & 
       \multicolumn{3}{|c|}{Overall $\pc$-F1}\\
       \hline
    & 2-Class & 3-Class  & 2-Class & 3-Class& 2-Class & 3-Class&  2-Class & 3-Class& 2-Class & 3-Class& 2-Class & 3-Class & 2-Class & 3-Class& 2-Class & 3-Class & LOCO\\
    \hline
    \textsc{Phrases+Roles}  & 0.675&0.675 & 0.420&0.420&  0.095 & 0.095 & 0.333 &0.333 & 0.449 &0.449&  0.234 &0.234& 0.385 &0.385 & 0.370 & 0.370 &0.370\\
    \textsc{GenAI+Roles} & 0.676 & 0.712&  0.535 & 0.600& 0.308 &0.727& 0.360 & 0.773 & 0.362 & 0.333  & 0.713 & 0.738 & 0.455 & 0.640& 0.487 & 0.646& 0.646\\
    SBert + SVM  & 0.944 & 0.944 & 0.582 & 0.578 & 0.000 & 0.083 & 0.432 & 0.427  &0.446& 0.483 &0.581 & 0.574 & 0.513& 0.509& 0.500 & 0.514& 0.419\\
    TFIDF + SVM& 0.935& 0.935&  0.679& 0.617 & 0.000 & 0.000 & 0.514 & 0.496 & 0.490 & 0.467 & 0.559 & 0.566 & 0.529 & 0.483  & 0.529 & 0.509 & 0.451 \\
     RoBERTa (fine-tuned) & 0.968 & 0.929&  0.736 & 0.717&  0.000 & 0.000&  0.565 & 0.642 & 0.550 & 0.587& 0.743& 0.811 &0.716& 0.723  & 0.611 & 0.630 & 0.610\\
    DistilBert (fine-tuned) & 0.956& 0.957 & 0.634& 0.644 & 0.000 & 0.000 & 0.527& 0.602 &0.533& 0.533& 0.774& 0.785 &0.795& 0.795  & 0.603 & 0.617& 0.582\\
    GPT4o & 0.730& 0.638 & 0.187 & 0.117  & 0.300 & 0.857 & 0.500& 0.515 &0.000& 0.000& 0.607& 0.629 &0.539& 0.348  & 0.409 & 0.443& 0.443\\
    \ourmethod & 0.946 & 0.929 
     & \textbf{0.755}$^*$ &\textbf{0.785}& 0.000& \textbf{1.000}$^*$ & \textbf{0.712}$^*$ & \textbf{0.894}$^*$ & \textbf{0.611} & \textbf{0.612}$^*$ & \textbf{0.811}$^*$ & \textbf{0.811}$^*$ & \textbf{0.811}$^*$ & \textbf{0.870}$^*$ & \textbf{0.663} & \textbf{0.843}$^*$&\textbf{0.688}$^*$\\
    \hline
    \end{tabular}
    }
    \caption{ 
      Here, we show the F1-score on the task of predicting \pc{} as there is class imbalance in all cities. Here we focus solely on inferring \pc{} (additional metrics shown in supplement). 
    The column LOCO contains the evaluations averaged across the set of cities in the 3-class setting, where each city is completely held out from the training once.
    In Richmond, less than 1\% of the test set utterances are \pc, leading to significant F1 variation. In the other cities, on average 10\% of the test set are \pc. 
    Results shown in bold indicate improvement over RoBERTa
    and those with an $*$ indicate statistically significantly improvement. }\label{quantresults}
\end{table*}

\subsubsection*{Comparators}
We construct two approaches which rely on domain knowledge of how meetings are typically structured. The first is \textsc{Phrases+Roles} and the second is \textsc{GenAI+Roles}. For each approach we first identify the discussion comment and hearing sections in a public meeting. Next, we use a series of rules to determine who within that section is speaking as a member of the public. We test two linguistic-based standards, an SVM with TF-IDF encodings and one with Sentence BERT embeddings. 
We also fine-tune two state-of-the-art comparators,  RoBERTa \citep{roberta} and DistilBERT \citep{distilbert} and GPT4o \cite{openai2024gpt4technicalreport}. The comparators are\footnote{More details on the baselines are in the supplement.}:

\begin{itemize}
    \item \textsc{Phrases+Roles} -  We first use public phrases to segment meetings into discussion sections. Next, we use \textsc{AssignRemarkTypes} to assign each utterance a remark type. 
    \item \textsc{GenAI+Roles}  - Like the previous approach we also use  \textsc{AssignRemarkTypes} to assign each utterance a remark type. However, we first segment meetings into discussion sections with \textsc{GenAISegments}. 
    \item SBert + SVM - This approach depends solely on the text content of meetings. We represent each utterance as an embedding using SentenceBert \citep{sbert}. Then we train an SVM on these representations. 
    \item TFIDF + SVM - We also train an SVM, but use a TFIDF encoding for the text. 
    \item  RoBERTa - We fine-tune RoBERTa on our annotated data, using the utterance text as input. 
    \item   DistilBERT  - We also fine-tune DistilBERT using the utterance text as input. 
    \item GPT4o - We use GPT4o as a zero-shot classifier by instructing it to detect \pc{} within a transcript.

\end{itemize}

\textsc{AssignRemarkTypes} looks at the sections during which a speaker is active, and the types of utterances they make to determine if they can best be described with a \textit{Public}
speaker role. Those utterances made during \comments{} or \hearings{} by those speaking as members of the public are labeled as $\pc$ or $\ph$ respectively. 

We initially experimented with different GenAI versions and prompts in order to obtain reasonable section predictions, on a small subset of data. This was done with a set of un-annotated meetings transcripts from across a range of cities. Finally, we used GPT-4.
Despite variations in how cities refer to public comment and hearing periods, and the topics allowed, each city typically introduces meeting sections in a structured manner. GenAI often detects this structure, even with varying specifications.
Here, our purpose was to test a general prompt, 
although if a team was interested in one city only, they could likely develop a prompt better customized to its structure.  The abbreviated prompt is below (the complete prompt is in the supplement):

\begin{quote}
    ``Public Comments'' typically but not always, allow the public to discuss non-agenda items, while ``Public Hearings'' are for comments on specific agenda items. Be aware that some meetings may not include these sections or might have multiple instances.
    To detect the start of these segments, look for somebody's call for comments. For identifying the end, focus on phrases indicating the conclusion of a section, such as  ``end'' and ``conclude''. 
\end{quote}

\subsubsection{Training Details}
Here, our goal is to mimic a real-world scenario. We assume we have access to 
some months of historical data. 
Given this data we would like to extract comments for meetings in the near future. 
Thus, 
we train on the first ${\sim}50$\% of each city's meetings (chronologically), 
validate on the next ${\sim}25$\%, and report results on the final ${\sim}25$\%. To ensure that our approach is not sensitive to the particular time period selected, 
we choose two different time periods for the quantitative evaluation, 2021, and 2023.\footnote{All experiments done on a server with a AMD EPYC 7763 64-Core Processor and 7 NVIDIA RTX A5000 GPUs. For the SentenceBert  model, we used the ``all-mpnet-base-v2'' language model.  For the RoBERTa model, we used ``roberta-large''.}

\subsubsection*{Results}
In  \tabref{quantresults} 
we see that \ourmethod{} performs the best 
in all three settings. 
The extent to which linguistic signals assist 
the overall predictions vary across cities. 
We see that in Royal Oak for example, the GenAI approach 
is quite effective, and may be useful when 
it is not possible to obtain annotations.
However, on average, incorporating linguistic signals
with PLMs is helpful. 
In Seattle, where public hearings are rare, 
and inter-annotator agreement is high, 
we see that detecting \pc\ is relatively straightforward
with PLMs alone. 
However, in every other city there are considerable gains 
from utilizing our structured approach. 
This is especially true when we pair our approach 
with our 3-class formulation. 
In  Oakland, Richmond, and Ann Arbor (cities with frequent public hearings), 
we see the strongest gains with this formulation and
 \ourmethod. 

\subsubsection{Generalizing to a New City}
In order to scale to cities across the US, our approach 
must be able to extract public comments for a previously unseen city. 
To evaluate our approach in the setting 
where a previously unseen city were to enter the dataset, 
we refactor the training data. 
Here, we keep the held-out test set the same for each city, 
but rather than train and validate a model on that city's data, 
we train and validate with a dataset that contains all of the training and validation 
data for all of the other cities. This allows us to test how well a model trained 
on one set of cities can generalize to a new unseen city. 
We refer to this in \tabref{quantresults} as Leave One City Out (LOCO).


\begin{figure*}[t] 
    \centering
    \begin{minipage}[b]{0.495\textwidth}
        \centering
        \includegraphics[scale=0.3055]{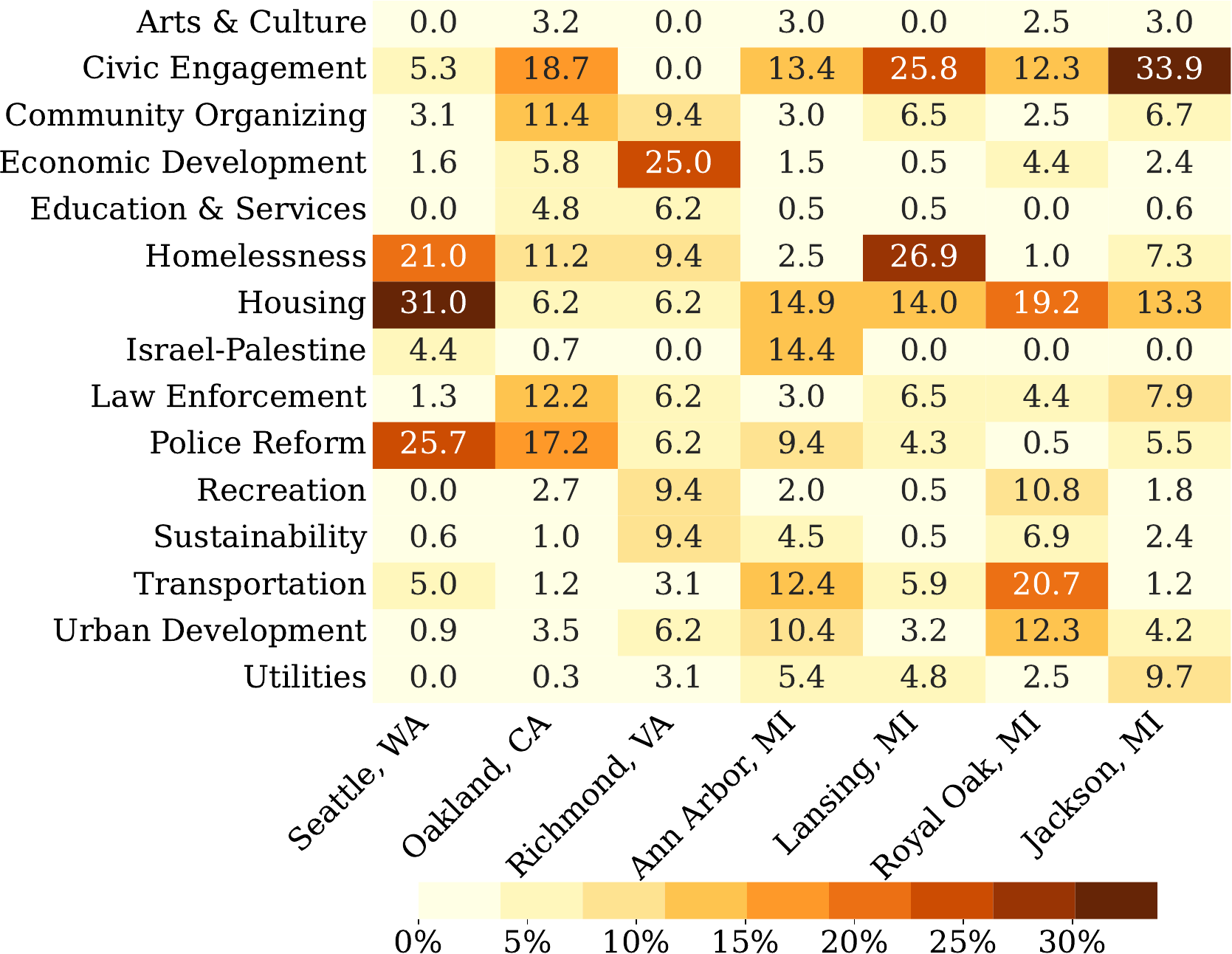}
    \end{minipage}
    \hfill
    \begin{minipage}[b]{0.495\textwidth}
        \centering
        \includegraphics[scale=0.3055]{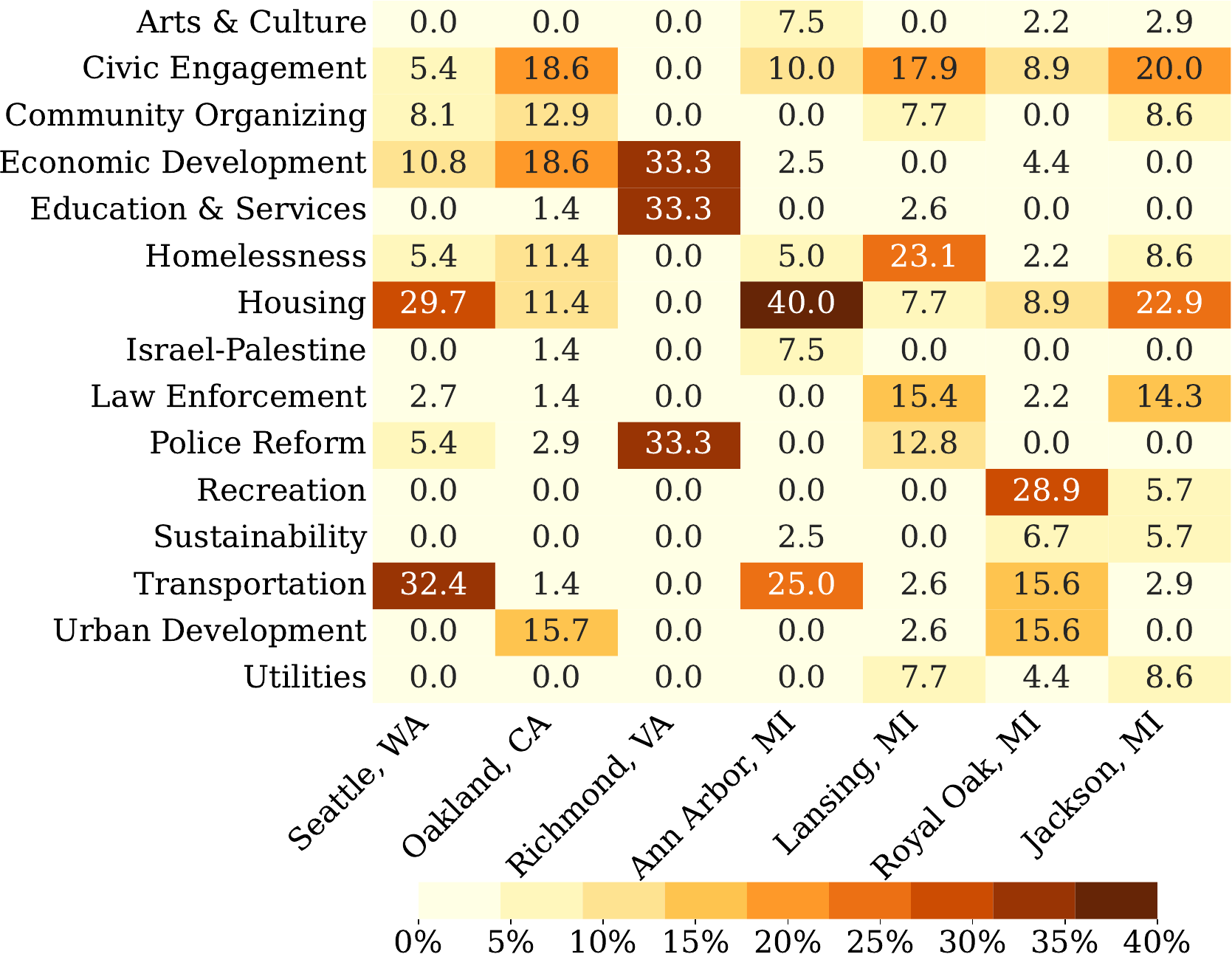}
    \end{minipage}
    
    \vspace{0.3cm}  
    \begin{minipage}[b]{0.60\textwidth}
        \centering
        \includegraphics[scale=0.3055]{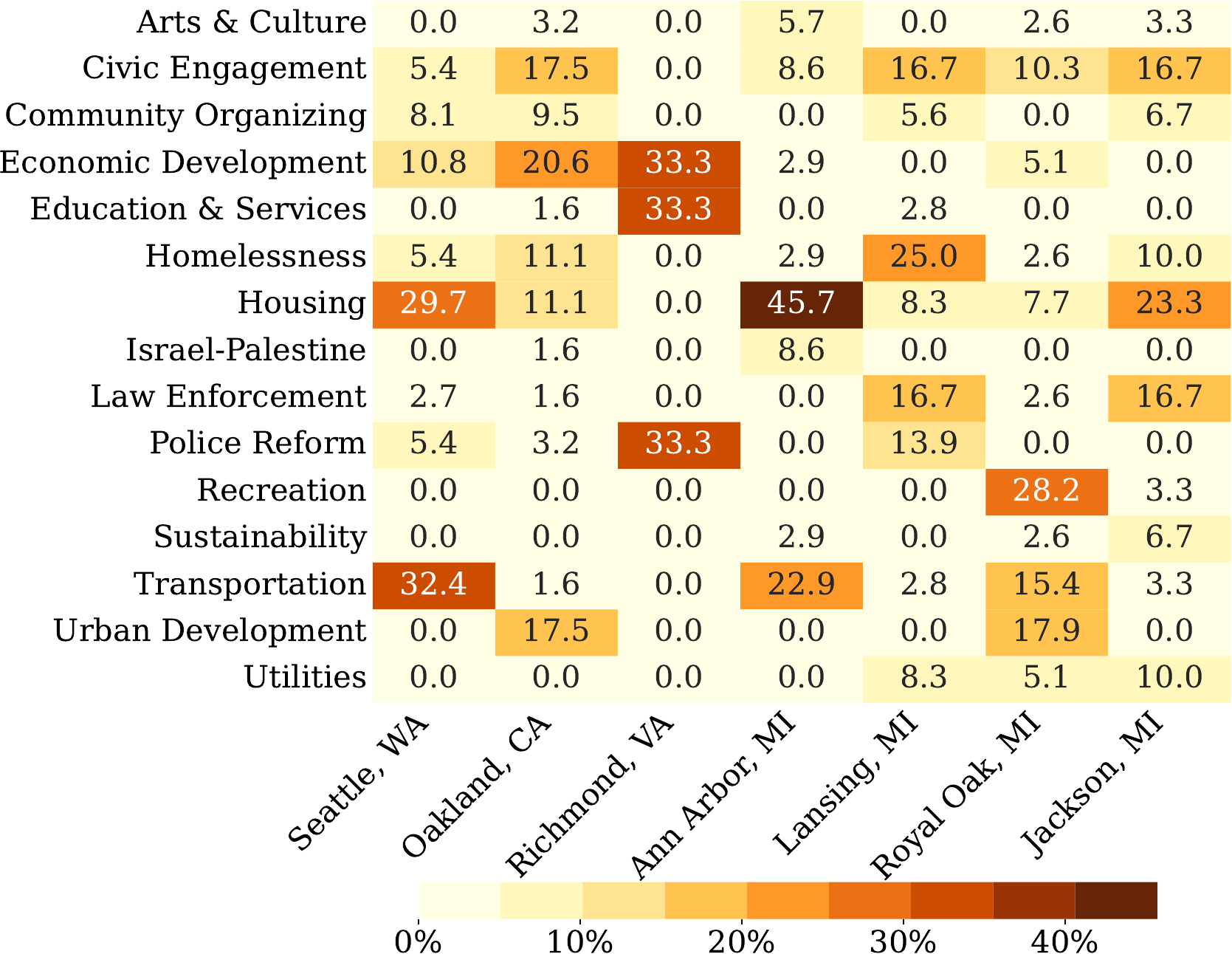}
    \end{minipage}

    \caption{Topics assigned to the public comments of each city. The top-left figure shows the topic assignments across pre-processed public comments in the training set for each city ($\sim$1700 total), 
    the top-right figure shows the topic assignments for those remarks in the test set which are known to be public comments ($\sim$350 total), and the bottom figure shows the topic assignments for those remarks in the test set which are predicted to be public comments by \ourmethod{} ($\sim$350 total). 
    We find that the top-right and bottom figures lead to qualitatively similar findings, indicating the efficacy of our approach.\footnote{Larger versions of all figures can be found in the supplement.}}
    \label{fig:topics}
\end{figure*}

\subsection{Hearing the Public}
The topics which are brought up by the public can 
provide a useful lens for comparing 
what different constituencies 
care about, 
and how these preferences might change over time
as residents respond to emerging issues. 
Here, we inspect the topics of public comments across all of the cities in the dataset. 
We experimented with a range of topic models before 
using a guided LLM approach (similar to  \cite{pham2023topicgpt}).
In our approach, we first use BERTopic \citep{grootendorst2022bertopic} 
to discover a category set of topics. We then refine this set and generate keywords 
for each topic with domain knowledge. Next, we combine each public comment with the potential topics and prompt an LLM (using Claude 3 Haiku \citep{claude}) to determine which topic best describes this remark. 
This  yielded the highest agreement with a manually assigned subset. 

In \figref{fig:topics}, we see that the topics of the comments made by the public differ 
greatly from city to city. 
 Housing, policing, and utilities (typical areas of responsibility for local government \citep{anzia2022local})
occur in almost all of the cities to various degrees. 
So do the topics of governance and civic engagement. 
Conversely, 
some topics are more localized. 
\textit{Police Reform} is strikingly relevant in Seattle and Oakland which both experienced large \textit{Defund the Police} movements following the murder of George Floyd in 2020.

The prevalence of the topic  \textit{Governance and Civic Engagement}, 
in the city of Jackson was surprising at first. 
However, this led us to discover that 
 Jackson held a contentious mayoral election in 2023. Opponents of the incumbent mayor frequently gave comments during public meetings as the general election approached. And while the  race received limited local news coverage, the high volume of public comments gave one early indication that the mayor's race would be a relatively close one.
 This leads us to believe that regularly running and updating topic models on public comments would
 lead to surprising insights for journalists. 

 Similarly, the inclusion of the \textit{Israel-Palestine} topic even before the recent conflict was somewhat surprising. 
 However, we found that in Seattle in 2021, public commenters voiced concerns about the Seattle Police Department receiving military training from the Israel Defense Forces \citep{dellacava2021}.
 In Ann Arbor, we found that there has been a regular movement for the city council to recognize injustice in Palestine
 which long predates October 7$^{th}$.


\section{Discussion}
We see that \ourmethod{} consistently outperforms the other methods, 
even in cities like Richmond with very few public comments per meeting. However there is variance across cities. 
In Seattle, a city with very few public hearings 
and where comment topics are often restricted, 
we see that PLMs achieve high performance 
and modeling structure is less advantageous.
The LLM does not achieve results of the same caliber. 
Conversely, in Ann Arbor and Oakland, 
where public comments can broach any topic
and there are often public hearings 
in council meetings, 
it is helpful to explicitly model meeting structure 
beyond linguistic signals, and to disambiguate remarks
made during \comments{} from those made during \hearings.
There is also better performance in the
3-class setting than in the 2-class setting. This is likely because explicitly modeling \ph{} allows for better disambiguation between \pc{} and \ph.

Our goal is to scale this pipeline across the United States. 
Towards this end, the ability of our approach to generalize
to a new city is a critical metric of success. 
Here, we see that our approach significantly beats the nearest competitor, 
RoBERTa by 13\%. 
We credit this with our ability to utilize structural signals.
While linguistic cues can be highly predictive of whether 
a remark is a public one, they can 
also vary across municipalities, limiting the ability to generalize to unseen cities from this information alone.
Our ability to apply domain knowledge and structural constraints 
makes our approach more robust to specific phrases which may be highly predictive in a given place or time
(e.g. a city-specific project such as the Oakland Coliseum project), 
but which are uninformative in a broader context.

\section{Contributions to Prior Work}
Our work contributes to efforts around 
improving government transparency. 
The Council Data Project \citep{brown2021council}
is an open-source civic technology project which aims to improve access to city council meetings, indexing and republishing meeting information to the web.
Civic Scraper is a
package which can be used to essential 
city council documents. 
LocalView \citep{barari2023localview}  compiles information on where to download public meeting records. 
Our contribution to these efforts is to provide an end-to-end
framework for producing a dataset of public remarks. 

We also contribute to a growing body of work on using technology to support journalism. 
At the extreme, some efforts may replace human-generated
content altogether. 
For example, automated journalism
is on the rise \citep{galily2018artificial,diakopoulos2019automating,clerwall2017enter,leppanen-etal-2017-data}.
Our goal is not to automate journalism,   
a path which may hold negative consequences \citep{waddell2018robot,latar2015robot,ferrara2023genai,illia2023ethical,kreps2022all,longoni2022news,jamil2021artificial}. 
Instead, our work is more closely aligned with 
efforts to use AI \textit{within} journalistic workflows. 
For example, the BBC has used chatbots to inform readers \citep{jones2021public}, while others have used bots to monitor news streams for misinformation \citep{mckinney2018tech,hassan2017toward,adair2017progress} and  fake news \citep{vo2018rise,alam2020fighting}.
These efforts and others \citep{marr2020tech, marshall2017bots}, aim to 
reduce the  time that newsrooms devote to menial tasks
\citep{van2012algorithms}

We contribute to the study of local politics by providing a novel method for studying political behavior at scale. This can add to existing studies of local government which have exposed disparities in who wields influence at the local level \citep{anzia2022local,einstein2019participates,einstein2020neighborhood}, and  highlighted the influence of interest groups in state politics \citep{grumbach2022laboratories}. 

Local government is a critical site of 
civic engagement and a participatory space for social justice. We 
contribute new data-driven insights on participation in
this democratic space, 
contributing to 
AI for Good \citep{tomavsev2020ai}.

\bibliography{aaai25}

\begin{thebibliography}{55}
\providecommand{\natexlab}[1]{#1}

\bibitem[{Achiam et~al.(2023)Achiam, Adler, Agarwal, Ahmad, Akkaya, Aleman, Almeida, Altenschmidt, Altman, Anadkat et~al.}]{openai2024gpt4technicalreport}
Achiam, J.; Adler, S.; Agarwal, S.; Ahmad, L.; Akkaya, I.; Aleman, F.~L.; Almeida, D.; Altenschmidt, J.; Altman, S.; Anadkat, S.; et~al. 2023.
\newblock GPT-4 Technical Report.
\newblock \emph{arXiv preprint arXiv:2303.08774}.

\bibitem[{Adair et~al.(2017)Adair, Li, Yang, and Yu}]{adair2017progress}
Adair, B.; Li, C.; Yang, J.; and Yu, C. 2017.
\newblock Progress Toward “the holy grail”: The Continued Quest to Automate Fact-Checking.
\newblock In \emph{Computation+ Journalism Symposium}.

\bibitem[{Alam et~al.(2021)Alam, Shaar, Dalvi, Sajjad, Nikolov, Mubarak, Da~San~Martino, Abdelali, Durrani, Darwish et~al.}]{alam2020fighting}
Alam, F.; Shaar, S.; Dalvi, F.; Sajjad, H.; Nikolov, A.; Mubarak, H.; Da~San~Martino, G.; Abdelali, A.; Durrani, N.; Darwish, K.; et~al. 2021.
\newblock Fighting the {COVID}-19 Infodemic: Modeling the Perspective of Journalists, Fact-Checkers, Social Media Platforms, Policy Makers, and the Society.
\newblock In \emph{Findings of the Association for Computational Linguistics (EMNLP)}, 611--649.

\bibitem[{{Anthropic}(2024)}]{claude}
{Anthropic}. 2024.
\newblock {The Claude 3 Model Family: Opus, Sonnet, Haiku}.

\bibitem[{Anzia(2022)}]{anzia2022local}
Anzia, S.~F. 2022.
\newblock \emph{Local Interests: Politics, Policy, and Interest Groups in US City Governments}.
\newblock University of Chicago Press.

\bibitem[{Bach et~al.(2017)Bach, Broecheler, Huang, and Getoor}]{bach2017hinge}
Bach, S.~H.; Broecheler, M.; Huang, B.; and Getoor, L. 2017.
\newblock Hinge-Loss Markov Random Fields and Probabilistic Soft Logic.
\newblock \emph{Journal of Machine Learning Research (JMLR)}, 18(109): 1--67.

\bibitem[{Bain et~al.(2023)Bain, Huh, Han, and Zisserman}]{bain2022whisperx}
Bain, M.; Huh, J.; Han, T.; and Zisserman, A. 2023.
\newblock WhisperX: Time-Accurate Speech Transcription of Long-Form Audio.
\newblock \emph{INTERSPEECH 2023}.

\bibitem[{Barari and Simko(2023)}]{barari2023localview}
Barari, S.; and Simko, T. 2023.
\newblock LocalView, A Database of Public Meetings for the Study of Local Politics and Policy-Making in the United States.
\newblock \emph{Scientific Data}, 10(1): 135.

\bibitem[{Brown et~al.(2021)Brown, Huynh, Na, Ledbetter, Ticehurst, Liu, Gilles, Cho, Ragoler, Weber et~al.}]{brown2021council}
Brown, E.~M.; Huynh, T.; Na, I.; Ledbetter, B.; Ticehurst, H.; Liu, S.; Gilles, E.; Cho, S.; Ragoler, S.; Weber, N.; et~al. 2021.
\newblock Council Data Project: Software for Municipal Data Collection, Analysis, and Publication.
\newblock \emph{Journal of Open Source Software (JOSS)}, 6(68): 3904.

\bibitem[{Brown and Weber(2022)}]{maxfield2022councils}
Brown, E.~M.; and Weber, N. 2022.
\newblock Councils in Action: Automating the Curation of Municipal Governance Data for Research.
\newblock \emph{Association for Information Science and Technology}, 59(1): 23--31.

\bibitem[{Brown and Tucker(2021)}]{brown2021philadelphia}
Brown, M.; and Tucker, E. 2021.
\newblock Philadelphia to Become First Major US City to Ban Police from Stopping Drivers for Low-level Traffic Violations.
\newblock \emph{Cable News Network}.

\bibitem[{CHS(2021)}]{dellacava2021}
CHS. 2021.
\newblock Sawant Says Will Pursue Legislation Banning Seattle Police Training with Israeli Military and Police.
\newblock \emph{Capitol Hill Seattle Blog}.

\bibitem[{Clerwall(2017)}]{clerwall2017enter}
Clerwall, C. 2017.
\newblock Enter the Robot Journalist: Users' Perceptions of Automated Content.
\newblock In \emph{The Future of Journalism: In an Age of Digital Media and Economic Uncertainty}, 165--177. Routledge.

\bibitem[{Diakopoulos(2019)}]{diakopoulos2019automating}
Diakopoulos, N. 2019.
\newblock \emph{Automating the News: How Algorithms Are Rewriting the Media}.
\newblock Harvard University Press.

\bibitem[{Einstein, Glick, and Palmer(2020)}]{einstein2020neighborhood}
Einstein, K.~L.; Glick, D.~M.; and Palmer, M. 2020.
\newblock Neighborhood defenders: Participatory politics and America’s housing crisis.
\newblock \emph{Political Science Quarterly (PSQ)}, 135(2): 281--312.

\bibitem[{Einstein, Palmer, and Glick(2019)}]{einstein2019participates}
Einstein, K.~L.; Palmer, M.; and Glick, D.~M. 2019.
\newblock Who Participates in Local Government? Evidence from Meeting Minutes.
\newblock \emph{Perspectives on Politics}, 17(1): 28--46.

\bibitem[{Ferrara(2024)}]{ferrara2023genai}
Ferrara, E. 2024.
\newblock GenAI against Humanity: Nefarious Applications of Henerative Artificial Intelligence and Large Language Models.
\newblock \emph{Journal of Computational Social Science (JCSS)}, 1--21.

\bibitem[{Galily(2018)}]{galily2018artificial}
Galily, Y. 2018.
\newblock Artificial Intelligence and Sports Journalism: Is It A Sweeping Change?
\newblock \emph{Technology in Society}, 54: 47--51.

\bibitem[{George(2008)}]{george2008internet}
George, L.~M. 2008.
\newblock The Internet and the Market for Daily Newspapers.
\newblock \emph{The BE Journal of Economic Analysis \& Policy (BEJEAP)}, 8(1).

\bibitem[{Greene(2024)}]{vpm2024}
Greene, B. 2024.
\newblock Richmond City Council Hears Public Comments on Ceasefire in Gaza.
\newblock \emph{VPM News}.

\bibitem[{Grootendorst(2022)}]{grootendorst2022bertopic}
Grootendorst, M. 2022.
\newblock BERTopic: Neural Topic Modeling with A Class-based TF-IDF Procedure.
\newblock \emph{arXiv preprint arXiv:2203.05794}.

\bibitem[{Grumbach(2022)}]{grumbach2022laboratories}
Grumbach, J. 2022.
\newblock \emph{Laboratories against Democracy: How National Parties Transformed State Politics}, volume 184.
\newblock Princeton University Press.

\bibitem[{Hanitzsch and Vos(2018)}]{hanitzsch2018journalism}
Hanitzsch, T.; and Vos, T.~P. 2018.
\newblock Journalism beyond Democracy: A New Look into Journalistic Roles in Political and Everyday Life.
\newblock \emph{Journalism}, 19(2): 146--164.

\bibitem[{Hassan et~al.(2017)Hassan, Arslan, Li, and Tremayne}]{hassan2017toward}
Hassan, N.; Arslan, F.; Li, C.; and Tremayne, M. 2017.
\newblock Toward Automated Fact-checking: Detecting Check-worthy Factual Claims by Claimbuster.
\newblock In \emph{the International Conference on Knowledge Discovery and Data Mining (KDD)}, 1803--1812.

\bibitem[{Illia, Colleoni, and Zyglidopoulos(2023)}]{illia2023ethical}
Illia, L.; Colleoni, E.; and Zyglidopoulos, S. 2023.
\newblock Ethical Implications of Text Generation in the Age of Artificial Intelligence.
\newblock \emph{Business Ethics, the Environment \& Responsibility (BEER)}, 32(1): 201--210.

\bibitem[{Jamil(2021)}]{jamil2021artificial}
Jamil, S. 2021.
\newblock Artificial Intelligence and Journalistic Practice: The Crossroads of Obstacles and Opportunities for the Pakistani Journalists.
\newblock \emph{Journalism Practice}, 15(10): 1400--1422.

\bibitem[{Jones and Jones(2021)}]{jones2021public}
Jones, B.; and Jones, R. 2021.
\newblock Public Service Chatbots: Automating Conversation with BBC News.
\newblock In \emph{Algorithms, Automation, and News}, 53--74. Routledge.

\bibitem[{Koenecke et~al.(2020)Koenecke, Nam, Lake, Nudell, Quartey, Mengesha, Toups, Rickford, Jurafsky, and Goel}]{koenecke2020racial}
Koenecke, A.; Nam, A.; Lake, E.; Nudell, J.; Quartey, M.; Mengesha, Z.; Toups, C.; Rickford, J.~R.; Jurafsky, D.; and Goel, S. 2020.
\newblock Racial Disparities in Automated Speech Recognition.
\newblock \emph{Proceedings of the National Academy of Sciences (PNAS)}, 117(14): 7684--7689.

\bibitem[{Kreps, McCain, and Brundage(2022)}]{kreps2022all}
Kreps, S.; McCain, R.~M.; and Brundage, M. 2022.
\newblock All the News that’s Fit to Fabricate: {AI}-generated Text as a Tool of Media Misinformation.
\newblock \emph{Journal of Experimental Political Science (JEPS)}, 9(1): 104--117.

\bibitem[{Latar(2015)}]{latar2015robot}
Latar, N.~L. 2015.
\newblock The Robot Journalist in the Age of Social Physics: The End of Human Journalism?
\newblock \emph{The New World of Transitioned Media: Digital Realignment and Industry Transformation}, 65--80.

\bibitem[{Lepp{\"a}nen et~al.(2017)Lepp{\"a}nen, Munezero, Granroth-Wilding, and Toivonen}]{leppanen-etal-2017-data}
Lepp{\"a}nen, L.; Munezero, M.; Granroth-Wilding, M.; and Toivonen, H. 2017.
\newblock Data-driven News Generation for Automated Journalism.
\newblock In \emph{the International Conference on Natural Language Generation (INLG)}, 188--197.

\bibitem[{Liu et~al.(2019)Liu, Ott, Goyal, Du, Joshi, Chen, Levy, Lewis, Zettlemoyer, and Stoyanov}]{roberta}
Liu, Y.; Ott, M.; Goyal, N.; Du, J.; Joshi, M.; Chen, D.; Levy, O.; Lewis, M.; Zettlemoyer, L.; and Stoyanov, V. 2019.
\newblock Roberta: A Robustly Optimized BERT Pretraining Approach.
\newblock \emph{arXiv preprint arXiv:1907.11692}.

\bibitem[{Longoni et~al.(2022)Longoni, Fradkin, Cian, and Pennycook}]{longoni2022news}
Longoni, C.; Fradkin, A.; Cian, L.; and Pennycook, G. 2022.
\newblock News from Generative Artificial Intelligence is Believed Less.
\newblock In \emph{the Conference on Fairness, Accountability, and Transparency (FAccT)}, 97--106.

\bibitem[{Marburger(2017)}]{marshall2017bots}
Marburger, J. 2017.
\newblock These Are the Bots Powering Jeff Bezos’ Washington Post Efforts to Build a Modern Digital Newspaper.
\newblock \emph{Nieman Lab}.

\bibitem[{Marr(2020)}]{marr2020tech}
Marr, B. 2020.
\newblock \emph{Tech Trends in Practice: The 25 Technologies that are Driving the 4th Industrial Revolution}.
\newblock John Wiley \& Sons.

\bibitem[{McKinney(2018)}]{mckinney2018tech}
McKinney, S. 2018.
\newblock Tech \& Check Alerts Aim to Ease the Workload of Fact-checkers.
\newblock \emph{Duke Reporters Lab}.

\bibitem[{Newman et~al.(2020)Newman, Fletcher, Schulz, And{\i}, and Nielsen}]{newman2020reuters}
Newman, N.; Fletcher, R.; Schulz, A.; And{\i}, S.; and Nielsen, R.-K. 2020.
\newblock Reuters {I}nstitute {D}igital {N}ews {R}eport 2020. {R}euters {I}nstitute. {U}niversity of {O}xford.

\bibitem[{Novick and Pickett(2022)}]{novick2022black}
Novick, R.; and Pickett, J.~T. 2022.
\newblock Black Lives Matter, Protest Policing, and Voter Support for Police Reform in {P}ortland, {O}regon.
\newblock \emph{Race and Justice (RAJ)}, 14(3): 368--392.

\bibitem[{Parasie and Dagiral(2013)}]{parasie2013data}
Parasie, S.; and Dagiral, E. 2013.
\newblock Data-driven Journalism and the Public Good: “Computer-assisted-reporters” and “Programmer-journalists” in Chicago.
\newblock \emph{New media \& society}, 15(6): 853--871.

\bibitem[{Pham et~al.(2023)Pham, Hoyle, Sun, and Iyyer}]{pham2023topicgpt}
Pham, C.~M.; Hoyle, A.; Sun, S.; and Iyyer, M. 2023.
\newblock TopicGPT: A Prompt-based Topic Modeling Framework.
\newblock \emph{arXiv preprint arXiv:2311.01449}.

\bibitem[{Phelps, Ward, and Frazier(2021)}]{phelps2021police}
Phelps, M.~S.; Ward, A.; and Frazier, D. 2021.
\newblock From Police Reform to Police Abolition? How Minneapolis Activists Fought to Make Black Lives Matter.
\newblock \emph{Mobilization: An International Quarterly}, 26(4): 421--441.

\bibitem[{Piotrowski and Borry(2010)}]{piotrowski2010analytic}
Piotrowski, S.~J.; and Borry, E. 2010.
\newblock An Analytic Framework for Open Meetings and Transparency.
\newblock \emph{Public Administration and Management}, 15(1): 138.

\bibitem[{Plaquet and Bredin(2023)}]{pyannote1}
Plaquet, A.; and Bredin, H. 2023.
\newblock {Powerset Multi-class Cross Entropy Loss for Neural Speaker Diarization}.
\newblock In \emph{INTERSPEECH 2023}.

\bibitem[{Powe(1992)}]{powe1992fourth}
Powe, L.~A. 1992.
\newblock \emph{The Fourth Estate and the Constitution: Freedom of the Press in America}.
\newblock University of California Press.

\bibitem[{Radford et~al.(2023)Radford, Kim, Xu, Brockman, McLeavey, and Sutskever}]{whisper}
Radford, A.; Kim, J.~W.; Xu, T.; Brockman, G.; McLeavey, C.; and Sutskever, I. 2023.
\newblock Robust Speech Recognition via Large-scale Weak Supervision.
\newblock In \emph{International Conference on Machine Learning (ICML)}, 28492--28518.

\bibitem[{Reimers and Gurevych(2019)}]{sbert}
Reimers, N.; and Gurevych, I. 2019.
\newblock Sentence-{BERT}: Sentence Embeddings using {S}iamese {BERT}-Networks.
\newblock In \emph{the Conference on Empirical Methods in Natural Language Processing (EMNLP)}.

\bibitem[{Sanh et~al.(2019)Sanh, Debut, Chaumond, and Wolf}]{distilbert}
Sanh, V.; Debut, L.; Chaumond, J.; and Wolf, T. 2019.
\newblock DistilBERT, A Distilled Version of BERT: Smaller, Faster, Cheaper and Lighter.
\newblock \emph{arXiv preprint arXiv:1910.01108}.

\bibitem[{Sullivan(2020)}]{sullivan2020ghosting}
Sullivan, M. 2020.
\newblock \emph{Ghosting the News: Local Journalism and the Crisis of American Democracy}.
\newblock JSTOR.

\bibitem[{Tomasev et~al.(2020)Tomasev, Cornebise, Hutter, Mohamed, Picciariello, Connelly, Belgrave, Ezer, Haert, Mugisha et~al.}]{tomavsev2020ai}
Tomasev, N.; Cornebise, J.; Hutter, F.; Mohamed, S.; Picciariello, A.; Connelly, B.; Belgrave, D.~C.; Ezer, D.; Haert, F. C. v.~d.; Mugisha, F.; et~al. 2020.
\newblock {AI} for Social Good: Unlocking the Opportunity for Positive Impact.
\newblock \emph{Nature Communications}, 11(1): 2468.

\bibitem[{Usher(2017)}]{usher2017appropriation}
Usher, N. 2017.
\newblock The Appropriation/Amplification Model of Citizen Journalism: An Account of Structural Limitations and the Political Economy of Participatory Content Creation.
\newblock \emph{Journalism Practice}, 11(2-3): 247--265.

\bibitem[{Van~Dalen(2012)}]{van2012algorithms}
Van~Dalen, A. 2012.
\newblock The Algorithms behind the Headlines: How Machine-written News Redefines the Core Skills of Human Journalists.
\newblock \emph{Journalism Practice}, 6(5-6): 648--658.

\bibitem[{Velazquez(2014)}]{velazquez2014solidarity}
Velazquez, M. 2014.
\newblock Solidarity and Empowerment in {C}hicago’s {P}uerto {R}ican print culture.
\newblock \emph{Latino Studies}, 12: 88--110.

\bibitem[{Vo and Lee(2018)}]{vo2018rise}
Vo, N.; and Lee, K. 2018.
\newblock The Rise of Guardians: Fact-checking URL Recommendation to Combat Fake News.
\newblock In \emph{the International Conference on Research \& Development in Information Retrieval (SIGIR)}, 275--284.

\bibitem[{Waddell(2018)}]{waddell2018robot}
Waddell, T.~F. 2018.
\newblock A Robot Wrote This? {H}ow Perceived Machine Authorship Affects News Credibility.
\newblock \emph{Digital Journalism}, 6(2): 236--255.

\bibitem[{Wilner, Montiel~Valle, and Masullo(2021)}]{wilner2021me}
Wilner, T.; Montiel~Valle, D.~A.; and Masullo, G.~M. 2021.
\newblock “{T}o Me, There’s Always a Bias”: Understanding the Public’s Folk Theories About Journalism.
\newblock \emph{Journalism Studies}, 22(14): 1930--1946.

\end{thebibliography}

\end{document}